\pdfoutput=1

\documentclass[11pt]{article}

\usepackage{acl}

\usepackage{times}
\usepackage{latexsym}

\usepackage[T1]{fontenc}

\usepackage[utf8]{inputenc}

\usepackage{microtype}
\usepackage{graphicx}
\usepackage{booktabs}
\usepackage{comment}
\usepackage[labelsep=quad,indention=10pt]{subfig}
\usepackage{bm}
\usepackage{wrapfig}

\usepackage{todonotes}
\usepackage[labelsep=quad,indention=10pt]{subfig}
\usepackage{multirow}
\usepackage{enumitem}
\usepackage{bm}
\usepackage{pifont}

\usepackage{amsfonts,amsmath,amssymb,amsthm}

\newcommand{\R}{\mathbb{R}}

\newcommand{\tr}[1]{#1^{\top}}
\newcommand{\cmark}{\ding{51}}%
\newcommand{\xmark}{\ding{55}}%

\newcommand{\red}[1]{\textcolor{red}{\bf #1}}
\newcommand\VisualSem{VisualSem}
\newcommand\VS{VS}

\title{Endowing Language Models with\\ Multimodal Knowledge Graph Representations}

\author{Ningyuan (Teresa) Huang$^1$\ \ ~~Yash R. Deshpande$^2$\ \  ~~Yibo Liu$^2$\ \ ~~Houda Alberts$^{3,}$\thanks{\ \ Work done while at the University of Amsterdam.}\\\bf Kyunghyun Cho$^2$\ \ Clara Vania$^{4,}$\thanks{\ \ Work done while at New York University.}\ \ Iacer Calixto$^{2,5}$\\\\
$^1$Johns Hopkins University, USA \ \ 
$^2$New York University, USA \\ 
$^3$Rond Consulting, NL \ \
$^4$Amazon Alexa AI, UK \\ 
$^5$Amsterdam UMC, University of Amsterdam, Dept. of Medical Informatics, Amsterdam, NL \\
\texttt{\{nh1724,yd1282,yl6769,ic1179\}@nyu.edu}
}

\date{}

\begin{document}
\maketitle
\begin{abstract}
We propose a method to make natural language understanding models more parameter efficient by storing knowledge in an external knowledge graph (KG) and retrieving from this KG using a dense index. Given (possibly multilingual) downstream task data, e.g., sentences in German, we retrieve entities from the KG and use their multimodal representations to improve downstream task performance.
We use the recently released \VisualSem{} KG as our external knowledge repository, which covers a subset of Wikipedia and WordNet entities, and compare a mix of tuple-based and graph-based algorithms to learn entity and relation representations that are grounded on the KG multimodal information.
We demonstrate the usefulness of the learned entity representations on two downstream tasks, and show improved performance on the multilingual named entity recognition task by $0.3\%$--$0.7\%$ F1, while we achieve up to $2.5\%$ improvement in accuracy on the visual sense disambiguation task.\footnote{All our code and data are available in: \url{https://github.com/iacercalixto/visualsem-kg}.}
\end{abstract}

\section{Introduction}

Recent natural language understanding (NLU) and generation (NLG) models 
obtain increasingly better state-of-the-art performance across benchmarks \citep{wang2018glue,nips2019superGLUE,xtreme2020}, however at the cost of a daunting increase in number of model parameters \citep{radford2019language,brown2020language,fedus2021switch}.\footnote{Although the OpenAI GPT-2, OpenAI GPT-3, and Google Switch models were released within a 2-years time span, they boast 1.5B, 175B, and 1.6T parameters, respectively.}
These increasingly large models lead to ever-increasing financial, computational, and environmental costs
\citep{strubell-etal-2019-energy}.
To make large language models (LMs) more parameter efficient, existing approaches propose to distill a compressed model given a large LM \citep{sanh2019distilbert} or to promote parameter sharing directly in the large LM \citep{Lan2020ALBERT}.

In this work, we advocate for an orthogonal approach:
to augment language models by retrieving multimodal representations from external structured knowledge repositories. In this way, LMs themselves need not store all this knowledge implicitly in their parameters \cite{petroni-etal-2019-language,nematzadeh-ruder-yogatama-2020-memory}.
Concretely, we investigate different methods to learn representations for entities and relations in the \VisualSem{} \citep[\VS{};][]{VisualSem2020} structured knowledge graph (KG).
We demonstrate the usefulness of the learned KG representations in multilingual named entity recognition (NER) and crosslingual visual verb sense disambiguation (VSD).
In particular, we use the \VS{} KG since it is a publicly available multilingual and multimodal KG designed to support vision and language research, 
and it provides a multimodal retrieval mechanism that allows neural models to retrieve entities from the KG using arbitrary queries.
In order to ground the downstream NER and VSD tasks with structured knowledge, we use \VisualSem{}'s \textit{sentence retrieval} model to map input sentences into entities in the KG, and use the learned entity representations as features.

\begin{figure*}[t!]
    \centering
    \includegraphics[width=0.8\textwidth]{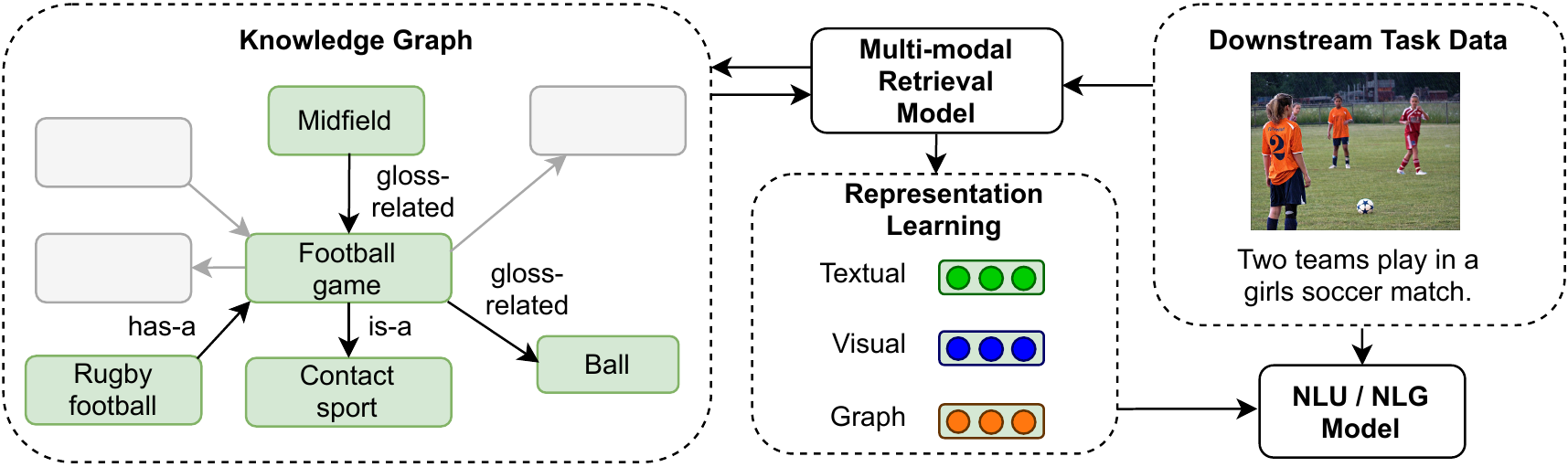}
    \caption{We learn visual, textual, and structural representations for a knowledge graph (KG). We propose to use this KG as an external knowledge repository, which we query using \textit{downstream task} sentences. We use the retrieved entities' representations to ground inputs in named entity recognition and visual verb sense disambiguation.}
    \label{fig:architecture_downstream}
\end{figure*}

We investigate different methods to learn representations for entities and relations in \VS{}, including tuple-based and graph-based neural algorithms.
We incorporate \VS{} textual descriptions in $14$ diverse languages as well as multiple images available for each entity in our representation learning framework.
Entity and relation representations are trained to be predictive of the knowledge graph structure, and we share them publicly with the research community to encourage future research in this area.
We illustrate our framework in Figure \ref{fig:architecture_downstream}.

Our main contributions are:
\begin{itemize}
    \item We compare the performance of different tuple- and graph-based representation learning algorithms for the \VisualSem{} KG.
    \item We demonstrate the usefulness of the learned representations on multilingual NER and crosslingual VSD, where we find that incorporating relevant retrieved entity representations improves model performance by up to $0.7\%$ F1 on NER and by up to $2.5\%$ accuracy in VSD.
    \item We publicly release our learned representations for entities and relations to encourage future research in this area.
\end{itemize}

This paper is organised as follows. In Section~\ref{sec:related} we discuss relevant related work, including solutions to endow models with access to external knowledge.
In Section~\ref{sec:visualsem} we briefly discuss \VisualSem{} as well as notation we use throughout the paper.
In Section~\ref{sec:models} we introduce the representation learning methods used in our experiments.
In Sections~\ref{sec:experimental_settings}, \ref{sec:results_no_additional_features} and \ref{sec:results_additional_features}, we introduce our experimental setup, 
evaluate and compare the learned representations with different architectures, and 
investigate the impact different modalities (i.e. textual and visual) have on the learned representations, respectively.
In Section \ref{sec:downstream-tasks}, we report experiments using the learned representations on downstream tasks.
Finally, in Section~\ref{sec:conclusions} we discuss our main findings as well as provide avenues for future work.

\section{Related work}
\label{sec:related}


\paragraph{Memory in Neural Models}
Endowing models with an external memory is not a new endeavour
\citep{Das92learningcontext-free,zhang-etal-1994discrete}. 
Approaches more directly relevant to our work include memory networks \citep[][]{weston2014memory,sukhbaatar2015end} and neural Turing machines \citep[][]{graves2014neural}, both proposed as learning frameworks that endow neural networks with an external memory that can be read from and written to.

Related to our \textit{multimodal} setting,
\citet{xiong2016dynamic} proposed memory networks for textual question answering (QA) and visual QA, where the memory module attends to the relevant visual features given the textual representations.
To address the issue where questions cannot be directly answered by the given visual content, \citet{su2018vqa} proposed visual knowledge memory networks that produce joint embedding of KG triplets and visual features.
\citet{wang2018m3} also applied a multimodal memory model ($M^3$) to the video captioning task. In all cases, memories are trained using task data and are used as a kind of \textit{working memory}  \citep{nematzadeh-ruder-yogatama-2020-memory}.

\paragraph{Retrieval augmented models}
In retrieval augmented models, memories are initialized with external knowledge beyond the data available for the task. 
\citet{lee-etal-2019-latent} proposed a framework for Open Retrieval QA (ORQA) where the retrieval and QA modules are jointly trained.
\citet{karpukhin2020dense} learned a dense passage retriever (DPR) and improved retrieval quality compared to standard sparse retrieval mechanisms, e.g., TF-IDF or BM25, which led to improvements on QA when using retrieved facts.
REALM \citep{guu2020realm} uses a dense Wikipedia index and fine-tunes the index together with a pretrained language model to address open-domain QA tasks.
\citet{petroni2020how} studied how feeding BERT with contexts retrieved/generated in different ways affects its performance on unsupervised QA without external knowledge.
Finally, \citet{lewis2020retrieval} proposed a retrieval augmented generation model where an encoder-decoder model learns to generate answers to questions conditioned on Wikipedia.

However, most existing retrieval augmented models based on Wikipedia do not usually include visual information or use
the structure in the KG. In this work, we propose a framework to retrieve multimodal information that encodes the structural information in the KG.

\paragraph{Multimodal pretraining}

Recently, pretrained vision and language models have achieved state-of-the-art results across many multimodal reasoning tasks \citep{tan2019lxmert, lu2019vilbert, yu2021ernievil}.
This line of work explicitly learns connections between vision and language inputs based on masked multimodal modelling over image-text pairs.
By contrast, we focus on modelling entities \textit{and relations} in an
\textit{entity-centric structured KG}, i.e., a KG where nodes denote concepts and contain multimodal information.

\section{\VisualSem{} Knowledge Graph}
\label{sec:visualsem}
\VisualSem{} \citep{VisualSem2020} is a multilingual and multimodal KG consisting of over $100$k nodes, $1.9$ million relations between nodes, and $1.5$ million associated images, built using BabelNet v4.0 \citep{navigli2012babelnet} and ImageNet \citep{ILSVRC2015}.
Each node denotes a \textit{synset} and is illustrated by multiple images.\footnote{A synset is a concept and can be described in many languages, e.g., synset \textit{dog} has associated description \textit{The dog is a mammal in the order Carnivora}.}
\VS{} covers many diverse topics and has $13$ relation types
with a strong visual component: \textit{is-a}, \textit{has-part}, \textit{related-to}, \textit{used-for}, \textit{used-by}, \textit{subject-of},
\textit{receives-action}, \textit{made-of}, \textit{has-property}, 
\textit{gloss-related}, \textit{synonym}, \textit{part-of}, and \textit{located-at}.
See \citet{VisualSem2020} for more details about the KG.

To the best of our knowledge, \VS{} is the only publicly available multimodal KG designed to be integrated into neural model pipelines, and thus chosen in our experiments.
Nonetheless, our framework can be applied to any KG.



\paragraph{Notation} Let the KG be a directed graph $\mathcal{G} = (\mathcal{V}, \mathcal{E})$, where $\mathcal{V}$ is the set of vertices or nodes, i.e., the KG \textit{synsets}, and $\mathcal{E}$ is the set of edges consisting of \textit{typed relations} between two synsets.
We refer to nodes $v_j$ connected to $v_i$ by any relation as part of $v_i$'s local neighborhood $\mathcal{N}_i$, i.e., $\forall v_j \in \mathcal{N}_i$, $v_j$ is related to $v_i$ by some relation $e_r$ in $\mathcal{E}$.

Let $\mathcal{D}$ be the set of all tuples ($v_i,e_r,v_j$) in the KG, i.e., all tuples where a \textit{head} node $v_i \in \mathcal{V}$ is related to a \textit{tail} node $v_j \in \mathcal{V}$ via a typed relation $e_r \in \mathcal{E}$ with \textit{type} $r$. Let $\mathcal{D}'$ be the set of corrupted tuples ($v_i,e_r,v'_j$), where ($v_i,e_r,v'_j$) is not in $\mathcal{G}$, and the tail node $v'_j$ (or head node $v'_i$) is randomly corrupted and not related to $v_i$ (or $v_j$) via $e_r$.
We learn representations for $\mathcal{G}$ including a node embedding matrix $\bm{V} \in \mathbb{R}^{|\mathcal{V}| \times d_n}$ and an edge embedding matrix $\bm{E} \in \mathbb{R}^{|\mathcal{E}| \times d_r}$, where $d_n$, $d_r$ are node and relation embedding dimensions, respectively.
Node $v_i$'s embedding $\bm{v}_i$ is the column vector $\bm{V}[i,:]$, and relation $e_r$'s embedding $\bm{e}_r$ is the column vector $\bm{E}[r,:]$. 
Each node $v_i \in \mathcal{V}$ is also associated to
a set of \textit{multilingual glosses} $\mathcal{T}_i$ and
\textit{images} $\mathcal{I}_i$.
Finally, we denote $[\bm{x}; \bm{y}]$ as the concatenation of $\bm{x}$ and $\bm{y}$, and $\bm{x} \odot \bm{y}$ as the element-wise multiplication.

\section{Knowledge Representation Learning}
\label{sec:models}

We are interested in learning robust multimodal representations for $\mathcal{V}$ and $\mathcal{E}$ that are useful in a wide range of downstream tasks. Next, we explore various tuple- and graph-based algorithms proposed to learn structured knowledge representations.


\subsection{Tuple-based algorithms}
Most tuple-based algorithms represent knowledge as a factual triple in the form of (\textit{head}, \textit{relation}, \textit{tail}) or using our previous notation, $(v_i,e_r,v_j)$, with its respective embeddings ($\bm{v}_i,\bm{e}_r,\bm{v}_j$).

\paragraph{TransE} 
TransE \citep{TransE} is a seminal neural KB embedding model.
It represents relations as translations in embedding space, where the embedding of the tail $\bm{v}_j$ is trained to be close to the embedding of the head $\bm{v}_i$ plus the relation vector $\bm{e}_r$, i.e., $\bm{v}_i + \bm{e}_r - \bm{v}_j \approx \bm{0}$.

\paragraph{DistMult}
The DistMult algorithm \citep{DistMult} is similar to TransE but uses a weighted element-wise dot product (\textit{multiplicative} operation) to combine two entity vectors.
The score for a triplet $(v_i,e_r,v_j)$ is computed as
\begin{align}
\phi(v_i,e_r,v_j)  &= \bm{v}_i \odot \bm{e}_r \odot \bm{v}_j. \label{eqn:DistMult}
\end{align}


\paragraph{TuckER} 
TuckER \citep{balazevic-etal-2019-tucker} models a KG as a three-dimensional tensor $\mathcal{X}\in \R^{|\mathcal{V}|\times|\mathcal{E}|\times|\mathcal{V}|}$,
and uses a Tucker decomposition \citep{tucker-decomposition-1966} to decompose $\mathcal{X}\approx\mathcal{W}\times_1\bm{V}\times_2\bm{E}\times_3\bm{V}$, with $\times_n$ indicating the tensor product along the n-th mode,
and $\mathcal{W}$ is the core tensor. 

The score function of a tuple is defined as $\phi(v_i, e_r,v_j)=\mathcal{W}\times_1\bm{v}_i\times_2 \bm{e}_r \times_3\bm{v}_j$.
Unlike TransE and DistMult which encode each tuple information directly into the relevant node ($\bm{V}$) and relation ($\bm{E}$) embeddings, TuckER stores shared information across different tuples in $\mathcal{W}$.

\subsection{Graph-based algorithms} 
Graph-based models compute node (and optionally edge) representations by learning an \textit{aggregation function} over node embeddings connected by its relations in the graph \citep{graph-neural-networks-2018}. 

\paragraph{GraphSage} 
\label{para:GraphSage}

The GraphSage \citep{GraphSage} algorithm 
computes node hidden states $\bm{h}_i$ by subsampling and
aggregating node $v_i$'s (subsampled) neighbors' states $\bm{h_j}, \forall v_j \in \mathcal{N}_i$.
We use GraphSage without subsampling and choose mean aggregation function, where the $l$-th layer hidden states are computed as:
\begin{equation}\label{eqn:graphsage}
\bm{h}_i^{(l+1)} = \texttt{ReLU} \left( \sum_{v_j \in \mathcal{N}_i} \frac{1}{| \mathcal{N}_i |}\bm{W}_s^{(l)} \bm{h}_j^{(l)}  \right)
\end{equation}
where $\bm{W}_s^{(l)}$ is a trained weight matrix.
In the first layer, $\bm{h}_j^{(l)}$ are set to the node embeddings $\bm{v}_j$. This is also similar to the formulation in graph convolutional networks (GCNs; \cite{kipf2016semi}). 

\paragraph{GAT}
\label{para:GAT}
In graph attention networks \citep[GAT;][]{GAT}, 
node hidden states  $\bm{h}^{(l+1)}_{i}$ are computed by aggregating neighboring nodes $\bm{h_j}, \forall v_j \in \mathcal{N}_i$, using an attention mechanism.
\begin{align}
\bm{h}_{i}^{(l+1)} &=\sigma\Big(\sum_{j \in \mathcal{N}_i} \alpha_{i j}^{(l)} \bm{z}_{j}^{(l)}\Big) , \label{eqn:gat1}\\
\bm{z}_{i}^{(l)} &=\bm{W}_g^{(l)} \bm{h}_{i}^{(l)} , \label{eqn:gat3}
\end{align}
where $\bm{W}_g^{(l)}$ is a matrix used to linearly project nodes states $\bm{h}^{(l)}_i$ and $\sigma$ is the sigmoid non-linearity. 
$\alpha_{i j}^{(l)}$ are attention scalar weights, computed as:
\begin{align}
\alpha_{i j}^{(l)} =\frac{\exp \left( \gamma\left( [ \bm{z}_{i}^{(l)}; \bm{z}_{j}^{(l)}] \cdot \bm{a}^{(l)} \right) \right)}{\sum_{k \in \mathcal{N}_i} \exp \left( \gamma\left( [ \bm{z}_{i}^{(l)}; \bm{z}_{k}^{(l)}] \cdot \bm{a}^{(l)} \right) \right)} , \label{eqn:gat2}
\end{align}
where $\bm{a}^{(l)}$ is a trained vector used to combine pairs of hidden states, and $\gamma$ is the $\operatorname{LeakyReLU}$ non-linearity. The overall mechanism is based on the additive attention mechanism introduced in \citet{Bahdanauetal2015Jointly}.

For simplicity, throughout the paper we denote the output of the last graph layer (GraphSage of GAT) for nodes $v_i \in \mathcal{V}$ to be $\bm{h}_i$ (without the layer superscript).

\subsubsection{Hybrid models}\label{sec:hybrid_models}

We also experiment with hybrid models where we combine
a graph-based model with a DistMult layer for prediction, since GraphSage and GAT do not normally use relation information in their message passing algorithms.
We are inspired by relational graph convolutional networks \citep{schlichtkrull2017modeling} where a DistMult layer was added to graph convolutional networks.
For a triplet $(v_i, e_r, v_j)$, given hidden states $\bm{h}_i$ and $\bm{h}_j$ generated from a graph-based model (GraphSage or GAT), we compute the triplet score in Equation \ref{eqn:DistMult} by replacing $\bm{v}_i$, $\bm{v}_j$ with $\bm{h}_i$, $\bm{h}_j$, respectively. 

\subsection{Multimodal Node Features}\label{sec:additional_features}
Each node $v_i \in \mathcal{V}$ includes two sets of additional features: textual features $\bm{t}_i$ for glosses $\mathcal{T}_i$, and visual features $\bm{m}_i$ for images $\mathcal{I}_i$.

\paragraph{Gloss Features}
Let $t_{i,g}$ be the $g$-th gloss in $\mathcal{T}_i$.
We extract features $\bm{t}_{i,g}$ for each gloss $t_{i,g} \in \mathcal{T}_i$ 
using Language-Agnostic Sentence Representations \cite[LASER;][]{LASER2019-TACL}, a 5-layer bidirectional LSTM model.
LASER supports 93 languages and obtains strong performance on cross-lingual tasks such as entailment and document classification. 
We tokenize glosses using the LASER tokenizer for the respective language.
Each gloss embedding $\bm{t}_{i,g} \in \mathcal{T}_i$ is a 1024-dimensional feature vector of max-pooled representations of the hidden states of the last bidirectional LSTM layer.

We aggregate text features $\bm{t}_{i}$ for each node $v_i \in \mathcal{V}$ as the average of its gloss features.
\begin{equation}
    \bm{t}_i = \frac{1}{|\mathcal{T}_i|} \sum_{g=1}^{|\mathcal{T}_i|}{\bm{t}_{i,g}} , \qquad \forall t_{i,g} \in \mathcal{T}_i.
\end{equation}
\noindent

\paragraph{Images Features}
Let $m_{i,l}$ be the $l$-th image in $\mathcal{I}_i$.
We extract features $\bm{m}_{i,l}$ for each image $m_{i,l} \in \mathcal{I}_i$ 
using a ResNet-152 \cite{he2016deep} architecture pretrained on ImageNet classification \cite{ILSVRC2015}.
We set each image embedding $\bm{m}_{i,l}$ to be the 2048-dimensional activation of the \textit{pool5} layer.

Similarly as we do for glosses, we aggregate visual features $\bm{m}_i$ for each node $v_i \in \mathcal{V}$ as the average of its image features.
\begin{equation}
    \bm{m}_i = \frac{1}{|\mathcal{I}_i|} \sum_{l=1}^{|\mathcal{I}_i|}{\bm{m}_{i,l}} , \qquad \forall m_{i,l} \in \mathcal{I}_i.
\end{equation}
\noindent

\subsubsection{Gating Text and Image Features}\label{sec:node-and-edge-gating}

\begin{figure}[t!]
    \centering
    \includegraphics[width=\linewidth]{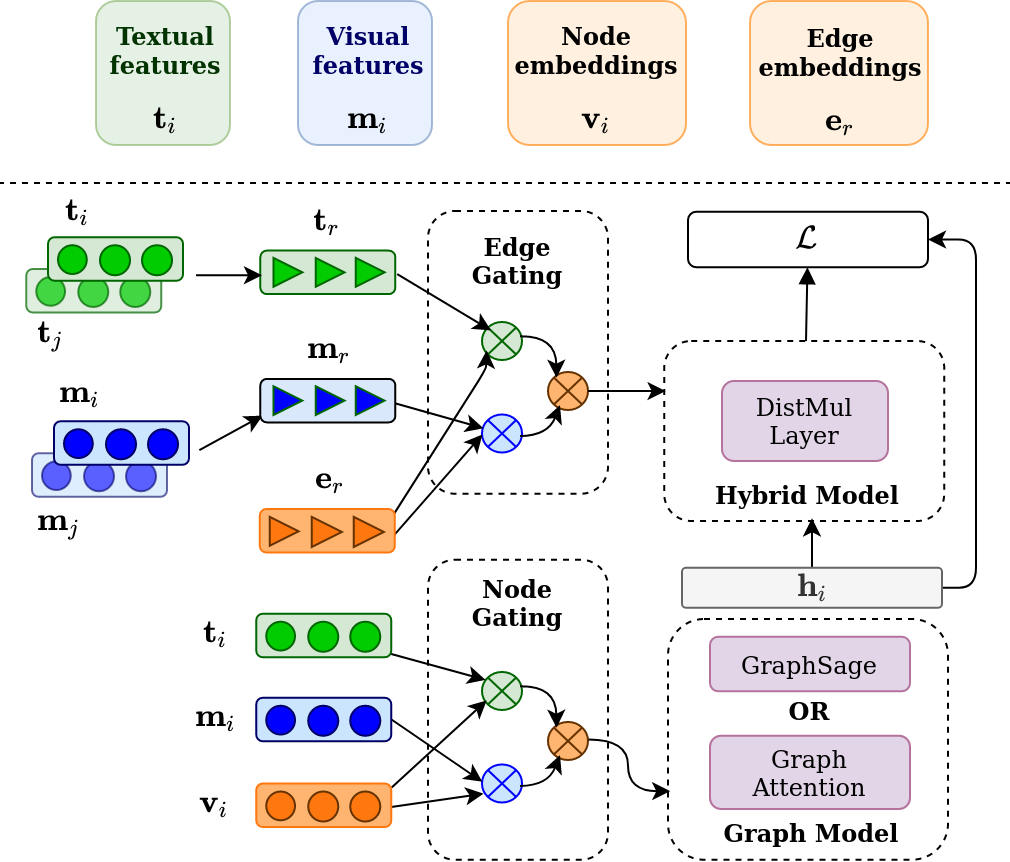}
    \caption{General architecture of our graph-based link prediction models. Node and edge embeddings $\bm{V}$ and $\bm{E}$ are trained and models optionally use textual and visual features $\bm{t}_i$, $\bm{m}_i$ for node $v_i$ to learn node and edge representations. Features are incorporated into node embeddings via node gating and into edge embeddings via edge gating, respectively.}
    \label{fig:architecture}
\end{figure}

We gate
gloss and image features $\{\bm{t}_i, \bm{m}_i\}$ in two different points in the model architecture: the node embeddings $\bm{v}_i$, before feeding it into graph layers (GraphSage or GAT), henceforth \textit{node gating};
and the edge hidden states $\bm{e}_r$, before feeding it to the DistMult layer, henceforth \textit{edge gating}.
Figure \ref{fig:architecture} illustrates the node and edge gating mechanisms.

\paragraph{Node Gating}
We denote the node gating function $f_n$ with trained parameters $\theta_n$ to transform node embeddings and additional features and compute \textit{informed} node embeddings.
For a given node embedding $\bm{v}_i$, we can integrate:
\textbf{1)} only textual features, in which case we have informed node embeddings $\bm{v}^t_i = f_n( \bm{v}_i, \bm{t}_i; \theta_n )$;
\textbf{2)} only image features, with informed node embeddings $\bm{v}^m_i = f_n( \bm{v}_i, \bm{m}_i; \theta_n )$; or
\textbf{3)} both textual and image features, with informed node embeddings $\bm{v}^{t,m}_i = f_n( \bm{v}_i, \bm{t}_i, \bm{m}_i; \theta_n )$.
Please see Appendix \ref{sec:appendix:node_gating} for details on the architectures we use to compute $f_n$ for (1), (2), and (3).

\paragraph{Edge (Relation) Gating}
Similarly to node gating, we denote the edge gating function $f_e$ with parameters $\theta_e$ to transform edge embeddings and additional features and compute \textit{informed} edge embeddings.
For a given edge embedding $\bm{e}_r$ denoting an edge between nodes $v_i$ and $v_j$, we integrate:
\textbf{1)} only textual features, in which case we have informed edge embeddings $\bm{e}^t_{r} = f_e( \bm{v}_i, \bm{t}_i, \bm{v}_j, \bm{t}_j; \theta_e )$;
\textbf{2)} only image features, with informed edge embeddings $\bm{e}^m_{r} = f_e( \bm{v}_i, \bm{m}_i, \bm{v}_j, \bm{m}_j; \theta_e )$; or
\textbf{3)} both textual and image features, with informed edge embeddings $\bm{e}^{t,m}_{r} = f_e( \bm{v}_i, \bm{t}_i, \bm{m}_i, \bm{v}_j, \bm{t}_j, \bm{m}_j; \theta_e )$.
Please see Appendix \ref{sec:appendix:edge_gating} for details on the architectures we use to compute $f_e$ for (1), (2), and (3).

\section{Experimental Setup}
\label{sec:experimental_settings}

We evaluate all models on the link prediction task, i.e., to identify whether there exists a \textit{relation} between a pair of \textit{head} and \textit{tail} nodes.
For each triplet ($v_i, e_r, v_j$) in the dataset, we create $k$ corrupted triplets 
where the tail $v_j$ (or the head $v_i$) is substituted by another node in the knowledge graph that is not connected with $v_i$ (or $v_j$) under relation $e_r$. 
We experiment with $k=\{100,1000\}$ corrupted examples per positive triplet.
See Appendix \ref{sec:appendix:training} for details on hybrid models GAT+DistMult and GraphSage+DistMult's architectures.

\paragraph{Training} All models are trained using negative sampling \cite{mikolov2013distributed} to maximize the probability of positive triplets while minimizing the probability of $k$ corrupted triplets.

\begin{equation}
\begin{split}
    \mathcal{L} = \frac{1}{| \mathcal{D} |} \sum_{(v_i,e_r,v_j) \in \mathcal{D}}
    \Big[- \log \sigma (\phi(v_i,e_r,v_j)) \\
  - \sum_{(v_i,e_r,v'_j) \in \mathcal{D}'} \log \sigma(- \phi(v_i,e_r,v'_j)) \Big],
\end{split}\label{eq:link_prediction_loss}
\end{equation}
where $\sigma(x)$ is the sigmoid function, $\mathcal{D}$ is the set of triplets in the knowledge graph $(v_i,e_r,v_j) \in \mathcal{G}$, $\mathcal{D}'$ is the set of corrupted triplets $(v_i,e_r,v'_j)$ $\notin \mathcal{G}$ .\footnote{For TransE and DistMult, we use both head-corrupted $(v'_i,e_r,v_j)$ and tail-corrupted triplets $(v_i,e_r,v'_j)$}

\paragraph{Scoring Function} Let $\phi(v_i,e_r,v_j)$ be the scoring function of a triplet.
For graph-based models, we compute the function as $\phi(v_i,e_r,v_j) = \tr{\bm{h}_i} \bm{h}_j$, i.e., there are no trained relation parameters.
For hybrid models, we compute $\phi(v_i,e_r,v_j) = \bm{h}_i \odot \bm{e}_r \odot \bm{h}_j$, i.e., we train the relation matrix $\bm{E}$.

When using multimodal features we replace the node embeddings input $\bm{v}_i$ with $\bm{v}_i^t$ when using text features, $\bm{v}^m_i$ when using image features, or $\bm{v}^{t,m}_i$ when using both.
For the hybrid models, in addition to the modified input, we also modify the input to the DistMult layer: instead of feeding $\bm{e}_r$, we use $\bm{e}_r^t$ for the text features, $\bm{e}_r^m$ for the image features, or $\bm{e}_r^{t,m}$ for both features.


\paragraph{Evaluation}
We follow the standard evaluation for link prediction using the mean reciprocal rank (MRR), i.e., the mean of the reciprocal rank of the correct triplet, and Hits@$\{1,3,10\}$, i.e., the proportion of correct triplets ranked in the top-$1$, top-$3$ or top-$10$ retrieved entries, respectively. The higher the number of corrupted tuples $k$ per positive example, the harder the task according to these metrics.
Finally, all results are an average over 5 different runs, and in each run models are selected according to the best validation MRR.

\paragraph{Hyperparameters}
We conduct an extensive hyperparameter search with the tuple-based and the graph-based models. See Appendix \ref{sec:appendix:hyperparameters} for details.

\begin{table}[t!]
    \centering
    \small
    \begin{tabular}{l rrrr}
    \toprule
    & \textbf{MRR} & \textbf{Hits@1} & \textbf{Hits@3} & \textbf{Hits@10} \\
    \midrule
    {TransE} & $3.2$ & $0.2$ & $3.3$ & $8.2$\\
    {DistMult} & $3.6$ & $1.9$ & $3.5$ & $7.6$  \\
    {Tucker} & $\bm{6.1}$ & $\bm{3.4}$ & $\bm{6.3}$ & $\bm{11.1}$ \\
    \bottomrule
    \end{tabular}
    \caption{Link prediction results on \VisualSem{}'s test set using all negative samples.}
    \label{tab:tuple-based-results-test-all-negative-samples}
\end{table}

\begin{table}[t!]
    \centering
    \small
    \resizebox{.48\textwidth}{!}{%
    \begin{tabular}{l l rrrr}
    \toprule
    & \bf R & \textbf{MRR} & \textbf{Hits@1} & \textbf{Hits@3} & \textbf{Hits@10} \\
    \midrule
    {TuckER} & \cmark & $19.0$ & $12.3$ & $17.7$ & $30.0$ \\
    {GAT} & \xmark & $10.0$ & $3.8$ & $12.6$ & $29.7$  \\
    \hspace{3pt} +{DistMult} & \cmark & $34.8$ & $13.6$ & $54.4$ & $69.3$ \\
    {GraphSage} & \xmark & $8.6$ & $2.3$ & $6.4$ & $18.0$ \\
    \hspace{3pt} +{DistMult} & \cmark & $\bm{78.4}$ & $\bm{56.8}$ & $\bm{100.0}$ & $\bm{100.0}$ \\
    \bottomrule
    \end{tabular}
    }
    \caption{Link prediction results on \VisualSem{}'s test set using 100 negative samples. {\bf R}: denotes whether the model learn relation features or not.}
    \label{tab:tuple-vs-graph-results-test-100-negative-samples}
\end{table}

\begin{table*}[t!]
    \centering
    \small
    \begin{tabular}{l rr rrrr rrrr}
    \toprule
    & \multicolumn{2}{c}{\textbf{Features}} & \multicolumn{4}{c}{\bf 100 negative examples} & \multicolumn{4}{c}{\bf 1000 negative examples} \\
    \cmidrule{4-11}
    & $\mathcal{T}_i$ & \bf $\mathcal{I}_i$ & \textbf{MRR} & \textbf{Hits@1} & \textbf{Hits@3} & \textbf{Hits@10} & \textbf{MRR} & \textbf{Hits@1} & \textbf{Hits@3} & \textbf{Hits@10} \\
    \midrule
    \multirow{3}{*}{\textbf{GAT}} & \xmark & \xmark & 34.8 & 13.6  & 54.4  & 69.3  & 4.4 & 0.0 & 0.0 & 4.7  \\
    \multirow{3}{*}{\textbf{+DistMult}} & \xmark & \cmark & 50.2 & 43.3  & 55.6  & 55.6  & $\underline{29.8}$  & 8.9  & 28.4  & 55.5   \\
    & \cmark & \xmark & $\underline{69.4}$ & $\underline{57.2}$  & $\underline{81.2}$  & $\underline{81.2}$  & 24.3  & 7.4  & 26.4  & $\underline{71.2}$   \\
    & \cmark & \cmark & 61.8 & 50.4  & 63.8  & 70.1 & 28.2  & $\underline{9.6}$  & $\underline{29.3}$  & 69.3  \\
    \cmidrule{2-11}
    \multirow{3}{*}{\textbf{GraphSage}} & \xmark & \xmark & 78.4 & 56.8  & $\bm{100.0}$  & $\bm{100.0}$  & 38.0  & 13.4  & 48.6  & $\bm{99.9}$   \\
    \multirow{3}{*}{\textbf{+DistMult}} & \xmark & \cmark & 80.7 & 61.5  & $\bm{100.0}$  & $\bm{100.0}$  & 46.9  & 31.9  & 47.2  & 98.3   \\
    & \cmark & \xmark & $\bm{84.7}$ & $\bm{69.5}$  & $\bm{100.0}$  & $\bm{100.0}$  & 36.4  & 13.8  & 42.8  & $\bm{99.9}$   \\
    & \cmark & \cmark & 80.7 & 61.4  & $\bm{100.0}$  & $\bm{100.0}$  & $\bm{61.6}$  & $\bm{50.6}$  & $\bm{63.6}$  & 97.2   \\
    \bottomrule
    \end{tabular}
    \caption{Link prediction results on \VisualSem{}'s test set with additional textual ($\mathcal{T}_i$) and visual features ($\mathcal{I}_i$). We show best overall scores per metric in bold, and underline best scores for a model across all features per metric.}
    \label{tab:graph-results-additional-features-test}
\end{table*}

\section{Results without Additional Features}
\label{sec:results_no_additional_features}

\subsection{Tuple-based}
We first investigate how tuple-based models fare when applied to link prediction on \VisualSem{}.
Tuple-based models are computationally more efficient than graph-based models, since they model the KB as a set of triplets. 
We train {\bf TransE} and {\bf DistMult} with OpenKE \citep{han2018openke}, and we train {\bf TuckER} following \citet{balazevic-etal-2019-tucker}.

Table
\ref{tab:tuple-based-results-test-all-negative-samples} reports each model performance on VisualSem's test set.
We present results when using all negative examples, i.e., using all other nodes in the knowledge graph as corrupted heads or tails given a triplet seen in the graph. We observe that TuckER outperforms both TransE and DistMult, with MRR and Hits@k almost twice as large as the other two algorithms. This is consistent with the finding reported by \citet{balazevic-etal-2019-tucker} on other knowledge graph datasets. Results on the validation set can be found in Appendix \ref{sec:appendix:link_prediction_no_features}.

\subsection{Tuple-based vs. Graph-based}
Given that TuckER is the best performing tuple-based model, we do not use TransE nor DistMult in further experiments and compare {\bf TuckER} to our graph-based models.
We first train vanilla {\bf GAT} and {\bf GraphSage} on link prediction with Deep Graph Library (DGL).\footnote{\url{https://www.dgl.ai/}} We also experiment with hybrid models where we add a DistMult layer to each graph-based model, thus {\bf GAT+DistMult} and {\bf GraphSage+DistMult}.

We show results in Table
\ref{tab:tuple-vs-graph-results-test-100-negative-samples}.
We highlight a few points: (1) graph-based baselines GAT and GraphSage perform poorly and are clearly outperformed by TuckER.
(2) However, TuckER uses relation features for prediction and is more comparable to hybrid models. 
Both GAT and GraphSage with a DistMult layer clearly outperform TuckER, which suggests that tuple-based models are worse than graph-based ones on link prediction in \VS{}, and learning relation features is crucial for a good performance.
We report results on the validation set in Appendix \ref{sec:appendix:link_prediction_no_features}.

\section{Results with Additional Features}
\label{sec:results_additional_features}

\paragraph{Graph-based}
In this set of experiments, we use our best performing graph-based model architectures and investigate the effect of using additional \textit{textual} and \textit{visual} features for the link prediction task.
We incorporate text features computed from the set of multilingual glosses $\mathcal{T}_i$ describing nodes $v_i \in \mathcal{V}$, and image features computed from the set of images $\mathcal{I}_i$ that illustrate $v_i$.
These features are integrated into the model architecture using node and edge gating modules described in Section \ref{sec:node-and-edge-gating}.

Table~\ref{tab:graph-results-additional-features-test} shows results for different graph-based representation learning models.
We first train and evaluate models using $k=100$ negative examples. We observe that incorporating both textual and visual features using GraphSage+DistMult yields the best link prediction performance.
However, we note that results obtained with GraphSage+DistMult achieves near-perfect scores for Hits@$k$ for $k>3$, which suggests that the number of negative examples is too small to observe meaningful differences from the additional features. 
Therefore, we also train and evaluate models using $k=1,000$ negative examples. We find a similar pattern and incorporating both features into GraphSage+DistMult yields the best overall performance. 
In general, we find that GAT+DistMult underperforms GraphSage+DistMult model. When using both textual and visual features, GraphSage+DistMult's performance is $33.4\%$ better according to MRR, and $33$--$40\%$ better according to Hits@$k$.
We report additional experiments on the validation set in Appendix \ref{sec:appendix:link_prediction_additional_features}.

\paragraph{Effects of Multimodal Features}
The overall impact of additional features is positive.
GAT+DistMult always improves with added features, although not consistently according to feature type. 
GraphSage+DistMult benefits the most when using both textual and visual features from the results with 1000 negative examples.  

\section{Evaluation on downstream tasks}\label{sec:downstream-tasks}

We use the representations learned with our best-performing models as additional features to ground a multilingual named entity recognition (\S \ref{sec:ner}) and a visual verb sense disambiguation model (\S \ref{sec:vsd}).

\paragraph{Node Representations}
We compute node hidden states $\bm{h}_i$ for all nodes $v_i \in \mathcal{V}$ with our best-performing model, GraphSage+DistMult.
We compare four different settings, which use node hidden states trained with: (1) no additional features ({\bf NODE}), (2) gloss features ({\bf TXT}), (3) image features  ({\bf IMG}), and (4) both gloss and image features ({\bf TXT+IMG}). We run a forward pass using each setting and generate 100-dimensional node hidden states: $\bm{h}_i^\text{\bf NODE}$,
$\bm{h}_i^\text{\bf TXT}$, 
$\bm{h}_i^\text{\bf IMG}$, 
$\bm{h}_i^\text{\bf TXT+IMG}$ (Sections \ref{sec:hybrid_models} and \ref{sec:additional_features}).
We select the best of the four settings according to validation set performance in each task, and use it to report test set results.

\subsection{Named Entity Recognition (NER)}
\label{sec:ner}

We use two NER datasets: \textbf{GermEval 2014} \citep{GermEval2014}, which contains data from German Wikipedia and news corpora,
and \textbf{WNUT-17} \citep{WNUT17}, an English NER dataset which includes user-generated text (e.g., social media, online forums, etc.).

\paragraph{Model} 
We use the pretrained English BERT and the multilingual BERT models for our experiments on WNUT-17 and GermEval, respectively.\footnote{We use HuggingFace's~\citep{Wolf2019HuggingFacesTS} models  \texttt{bert-large-cased} and \texttt{bert-base-multilingual-cased} available in \url{https://huggingface.co/}.}
For our baseline, we simply fine-tune BERT on the NER data using a multi-layer perceptron (MLP) classifier after the final BERT layer. Let $\bm{z}_i$ be the final layer representation for an input word $x_i$.
The probabilities of the correct label are:
\begin{align}
    \bm{\hat{y}}_i &= \texttt{softmax} (\bm{W}^n \bm{z}_i),
    \label{eq:ner-model1}
\end{align}
where $\bm{W}^n$ is the classification head.
To include additional features from \VisualSem{}, we use \VisualSem{}'s sentence retrieval model to retrieve the top-$k$ closest nodes in the \VisualSem{} graph 
for each input sentence.

We experiment with two strategies:
\textbf{i) concat}, where we first concatenate $\bm{z}_i$ with the representation of the top-$1$ retrieved node $\bm{h}_i^{RET}$ and then predict the label as below.
\begin{align}
    \bm{\hat{y}}_i &= \texttt{softmax} (\bm{W}^n [\bm{z}_i;\bm{W}^{RET}\bm{h}_i^{RET}]),
    \label{eq:ner-model2}
\end{align}
and \textbf{ii) attend}, where we use an attention mechanism to summarize the top-$5$ retrieved nodes using $\bm{z}_i$ as the query, and then concatenate the returned vector to the original hidden state $\bm{z}_i$:
\begin{align}
    \bm{\hat{y}}_i &= \texttt{softmax} (\bm{W}^n [\bm{z}_i;\bm{W}^{RET} \cdot \bm{a}]),\\
    \bm{a} &= \texttt{Attention} (\bm{z}_i, \{\bm{h}_i^{RET} \}_{k=1}^5).
    \label{eq:ner-model3}
\end{align}


\paragraph{Results}
Table \ref{tab:NER-test-results} shows our results on test sets. Adding \VisualSem{} entity representations trained with node features yields moderate improvements of $0.7\%$ F1 compared to the baseline for the WNUT-17 (\textbf{EN}) dataset, and $0.3\%$ on the GermEval (\textbf{DE}) dataset.
In both cases, the best results are obtained when adding entity representations trained using \VisualSem{}'s graph structure only (i.e., not using any textual or visual information).

\begin{table}[t!]
    \centering
    \small
    \resizebox{\columnwidth}{!}{%
    \begin{tabular}{@{}l l@{} ccc@{}}
    \toprule
    & &  \textbf{Precision} & \textbf{Recall} & \textbf{F1 Score} \\
    \midrule
     \parbox[t]{2mm}{\multirow{3}{*}{\rotatebox[origin=c]{90}{\textbf{EN}}}}
     &{\textbf{Baseline}} & $58.4$ & $\bm{39.9}$ & $47.4$ \\
    & \hspace{5pt}\textbf{+concat} $\bm{h}_i^\text{\bf IMG}$  &$57.1$ & $39.1$ & $46.4$ \\
    & \hspace{5pt}\textbf{+attend} $\{\bm{h}_i^\text{\bf NODE}\}_{k=1}^5$  & $\bm{61.5}$ & $39.5$ & $\bm{48.1}$ \\
    \midrule
    \parbox[t]{2mm}{\multirow{3}{*}{\rotatebox[origin=c]{90}{\textbf{DE}}}}
    &{\textbf{Baseline}} & 86.0          & 86.2          & 86.1          \\
    & \hspace{5pt}\textbf{+concat} $\bm{h}_i^\text{\bf NODE}$  &$\bm{86.2}$ & $\bm{86.6}$ & $\bm{86.4}$ \\
    & \hspace{5pt}\textbf{+attend} $\{\bm{h}_i^\text{\bf TXT+IMG}\}_{k=1}^5$  & $85.7$ & $86.0$ & $85.9$ \\
    \bottomrule
    \end{tabular}
    }
    \caption{NER results on the WNUT-17 (\textbf{EN}) and GermEval (\textbf{DE}) test sets.}
    \label{tab:NER-test-results}
\end{table}

\subsection{Crosslingual Visual Verb Sense Disambiguation}
\label{sec:vsd}

Crosslingual visual verb sense disambiguation (VSD) is the task of choosing the correct translation given an ambiguous verb in the source language, i.e., a verb with more than one possible translation.



We use \textbf{MultiSense} \citep{gella2019crosslingual}, a collection of 9,504 images covering 55 English verbs with their unique translations into German and Spanish. Each image is annotated with a translation-ambiguous English verb, a textual description, and the correct translation in the target language.\footnote{\url{github.com/spandanagella/multisense/}} In our experiments, we use the German verb translations of the dataset, which consists of 154 unique German verbs. 

\paragraph{Model}
We encode the $i$-th image using the pretrained ResNet-152 \citep{he2016deep} convolutional neural network and use the \textit{pool5} 2048-dimensional activation as the visual features $\bm{z}^m_i$.
We encode the $i$-th English verb and its textual description (separated by \texttt{[SEP]} token) using a pretrained BERT model\footnote{We use the \texttt{bert-large-cased} model.} and use the final layer English verb token representation as the textual features $\bm{z}^t_i$. We do not fine-tune the BERT parameters and only train the task's classifier head.

To predict the target translation, we concatenate the projected textual and visual features, pass the resulting tensor through a 128-dimensional hidden layer followed by a non-linearity. We then apply another projection layer, followed by a softmax.
\begin{align}
    \bm{h}_i &= \texttt{ReLU}(\bm{W}^h[\bm{W}^m \bm{z}^m_i; \bm{W}^t \bm{z}^t_i]),  \label{eqn:multi1}\\ 
    \bm{\hat{y}}_i &= \texttt{softmax} (\bm{W}^o \bm{h}_i).\label{eqn:multisense}
\end{align}



To incorporate information from the knowledge graph, we retrieve the top-1 nearest node representation $\bm{h}^{RET}_i$ for the $i$-th input, which is the concatenation of the English verb and the textual description. 
We then concatenate $\bm{h}^{RET}_i$ to the input, and instead compute $\bm{h}_i$ as below.
\begin{equation}
    \bm{h}_i = \texttt{ReLU}(\bm{W}^h [\bm{W}^m \bm{z}^m_i; \bm{W}^t \bm{z}^t_i; \bm{h}^{RET}_i]),  \label{eqn:multi2}
\end{equation}
where the hidden layer size is reduced to 100 such that the number of trained parameters in \eqref{eqn:multi1} and \eqref{eqn:multi2} is comparable.

\paragraph{Results} 
Table \ref{tab:Multisense-test-results} shows our results. 
Overall, adding \VisualSem{} representations generated with image features yields the best performance compared to the baseline without additional multimodal features. This suggests that crosslingual visual VSD benefits from the additional multimodal information encoded with structural and visual context.


\begin{table}[t!]
    \centering
    \small
    \begin{tabular}{lr}
    \toprule
    & \textbf{Accuracy} \\
    \midrule
      \citep{gella2019crosslingual} & 55.6 \\
      {\textbf{Our Baseline}} & 94.4 \\
      \quad \textbf{+}$\bm{h}_i^\text{\bf NODE}$ & 96.8 \\
      \quad \textbf{+}$\bm{h}_i^\text{\bf IMG}$ & \textbf{97.2} \\
    \bottomrule
    \end{tabular}
    \caption{Accuracy on the MultiSense test set (German).}
    \label{tab:Multisense-test-results}
\end{table}

\section{Conclusions and Future work}
\label{sec:conclusions}

We conducted a systematic comparison of different tuple- and graph-based architectures to learn robust multimodal representations for the \VisualSem{} knowledge graph.
We found that additionally using visual and textual information available to nodes in the graph (illustrative images and descriptive glosses, respectively) leads to better node and entity representations, as evaluated on link prediction.
Using our best-performing method, a hybrid between a tuple- and graph-based algorithm, we demonstrated the usefulness of our learned entity representations in two downstream tasks:
we obtained substantial improvements on crosslingual visual VSD by $2.5\%$ accuracy 
compared to a strong baseline, and
we moderately improved performance for multilingual NER between 0.3\%--0.7\% F1.
In both cases, we used simple downstream architectures and could likely improve results even further by carefully experimenting with different integration strategies.

We envision many possible avenues for future work.
We will incorporate more downstream tasks in our evaluation, including vision-focused tasks (e.g., object detection) and challenging generative tasks (e.g., image captioning).
We also plan to apply our framework to other KGs to integrate different types of information, e.g., commonsense knowledge from ConceptNet.
Finally, another possibility is to include structured KG representations in large retrieval-based LMs.

\section*{Acknowledgments}
IC has received funding from the European Union’s Horizon 2020 research and innovation programme under the Marie Skłodowska-Curie grant agreement No 838188. NH is partially supported by the Johns Hopkins Mathematical Institute for Data Science (MINDS) Data Science Fellowship.

CV's work on this project at New York University was financially supported by Eric and Wendy Schmidt (made by recommendation of the Schmidt Futures program) and Samsung Advanced Institute of Technology (under the project \textit{Next Generation Deep Learning: From Pattern Recognition to AI}) and benefitted from in-kind support by the NYU High-Performance Computing Center. This material is based upon work supported by the National Science Foundation under Grant No. 1922658. Any opinions, findings, and conclusions or recommendations expressed in this material are those of the author(s) and do not necessarily reflect the views of the National Science Foundation.

\bibliography{anthology,refs}
\bibliographystyle{acl_natbib}

\clearpage
\appendix

\section{Hyperparameters}\label{sec:appendix:hyperparameters}
\paragraph{TransE} We perform a grid search over the learning rate $\alpha \in \{0.01, 0.1, 1\}$, embedding dimension $d \in \{100, 200, 300, 400\}$, and margin $m \in \{1, 5, 10\}$. The optimal configuration is $\alpha = 1$, $d = 200$, and $m = 5$. In training, the number of negative examples is fixed to 25.

\paragraph{DistMult}
We perform a grid search over the learning rate $\alpha \in \{0.1, 0.01, 0.001\}$, embedding dimension $\in \{100, 200\}$, and weight decay rate $\lambda \in \{0.01, 0.0001\}$. The optimal configuration is $\alpha = 0.1$, $d = 100$, and $\lambda = 0.0001$.

\paragraph{TuckER}
We conduct a grid search over the learning rate $\alpha \in \{0.01, 0.005, 0.003, 0.001, 0.0005\}$ and
weight decay rate $\lambda \in \{1, 0.995, 0.99\}$.
Entity and relation embedding dimensions are fixed at $200$ and $30$ respectively.
The optimal configuration is $\alpha = 0.03$, $\lambda = 1$, input dropout $= 0.2$, dropout $= 0.2$, and label smoothing rate $= 0.1$.

\paragraph{GraphSage and GraphSage+DistMult}
Both models are trained for a maximum of 100 epochs and we early stop if validation MRR does not improve for over 20 epochs. We search for learning rate
$\{0.001, 0.002\}$ 
and weight decay $\{0, 0.0001, 0.00005, 0.00001\}$.
We use $100$-dimensional node and relation embeddings, and two graph layers (Equation \ref{eqn:graphsage}) with a dropout rate of $0.5$ applied to the input of each layer.
Both models have batch size $10^4$ for $100$ negative examples and $10^3$ for $1000$ negative examples.

\paragraph{GAT and GAT+DistMult} Both models are trained for a maximum of $100$ epochs, and we also early stop if validation MRR does not improve for over $20$ epochs. We perform a grid search over the learning rate $\alpha \in \{0.1, 0.001, 0.0001\}$, the embedding dimension $d \in \{100, 200\}$, number of attention heads $n_{h} \in \{2, 4\}$, and weight decay rate $\lambda \in \{0, 0.001, 0.005, 0.0001\}$. The optimal configuration is $\alpha = 0.001$, $d = 100$, $n_{h} = 2$, and $\lambda = 0.005$.
Both models have batch size $10^4$ for $100$ negative examples and $10^3$ for $1000$ negative examples.

\section{Link Prediction Results}\label{sec:appendix:link_prediction}

\subsection{Without Additional Features}\label{sec:appendix:link_prediction_no_features}
In Tables \ref{tab:tuple-based-results-validation-all-negative-samples} and \ref{tab:tuple-vs-graph-results-validation-100-negative-samples} we show additional results on  \VisualSem{} link prediction  on the validation set, obtained with tuple-based, graph-based, and hybrid models.
We note that results are similar to the ones obtained on the test set and reported in Tables \ref{tab:tuple-based-results-test-all-negative-samples} and \ref{tab:tuple-vs-graph-results-test-100-negative-samples}.

\begin{table}[t!]
    \centering
    \small
    \begin{tabular}{l rrrr}
    \toprule
    & \textbf{MRR} & \textbf{Hits@1} & \textbf{Hits@3} & \textbf{Hits@10} \\
    \midrule
    {TransE} & $3.5$ & $0.3$ & $3.7$ & $9.3$ \\
    {DistMult} & $3.7$ & $1.9$ & $3.6$ & $7.7$ \\
    {Tucker} & $\bm{6.2}$ & $\bm{3.5}$ & $\bm{6.2}$ & $\bm{11.3}$ \\
    \bottomrule
    \end{tabular}
    \caption{Link prediction results on \VisualSem{}'s validation set using all negative samples.}
    \label{tab:tuple-based-results-validation-all-negative-samples}
\end{table}

\begin{table}[t!]
    \centering
    \small
    \resizebox{.48\textwidth}{!}{%
    \begin{tabular}{l l rrrr}
    \toprule
    & \bf R & \textbf{MRR} & \textbf{Hits@1} & \textbf{Hits@3} & \textbf{Hits@10} \\
    \midrule
    {TuckER} & \cmark & $58.7$ & $47.5$ & $64.4$ & $80.3$ \\
    {GAT} & \xmark & $10.0$ & $4.0$ & $12.7$ & $29.8$ \\
    \hspace{3pt} +{DistMult} & \cmark & $\bm{75.0}$ & $\bm{61.2}$ & $88.6$ & $88.6$ \\
    {GraphSage} & \xmark & $8.7$ & $2.4$ & $6.5$ & $18.1$ \\
    \hspace{3pt} +{DistMult} & \cmark & $74.9$ & $49.8$ & $\bm{100.0}$ & $\bm{100.0}$ \\
    \bottomrule
    \end{tabular}
    }
    \caption{Link prediction results on \VisualSem{}'s validation set using 100 negative samples. {\bf R}: denotes whether the model learn relation features or not.}
    \label{tab:tuple-vs-graph-results-validation-100-negative-samples}
\end{table}

\subsection{Additional Features}\label{sec:appendix:link_prediction_additional_features}
We report link prediction results on \VisualSem{}'s validation set when models have access to additional features (i.e., gloss and/or image features) in Table \ref{tab:graph-results-additional-features-validation}.

\begin{table*}[t!]
    \centering
    \small
    \begin{tabular}{l rr rrrr rrrr}
    \toprule
    & \multicolumn{2}{c}{\textbf{Features}} & \multicolumn{4}{c}{\bf 100 negative examples} & \multicolumn{4}{c}{\bf 1000 negative examples} \\
    \cmidrule{4-11}
    & $\mathcal{T}_i$ & \bf $\mathcal{I}_i$ & \textbf{MRR} & \textbf{Hits@1} & \textbf{Hits@3} & \textbf{Hits@10} & \textbf{MRR} & \textbf{Hits@1} & \textbf{Hits@3} & \textbf{Hits@10} \\
    \midrule
    \multirow{3}{*}{\textbf{GAT}} & \xmark & \xmark & $75.0$ & $61.2$ & $88.6$ & $88.6$ & $\underline{34.2}$ & $\underline{14.6}$ & $\underline{41.0}$ & $\underline{87.9}$ \\
    \multirow{3}{*}{\textbf{+DistMult}} & \xmark & \cmark & $\underline{79.6}$ & $\underline{66.6}$ & $\underline{92.5}$ & $\underline{92.5}$ & $30.8$ & $11.8$ & $36.6$ & $83.4$ \\
    & \cmark & \xmark & $71.4$ & $59.0$ & $83.6$ & $83.6$ & $28.2$ & $9.1$ & $32.9$ & $83.4$ \\
    & \cmark & \cmark & $70.9$ & $57.9$ & $83.6$ & $83.6$ & $30.5$ & $11.7$ & $35.0$ & $83.4$ \\
    \cmidrule{2-11}
    \multirow{3}{*}{\textbf{GraphSage}} & \xmark & \xmark & $74.9$ & $49.8$ & $\bm{100.0}$ & $\bm{100.0}$ & $39.6$ & $19.0$ & $43.7$ & $\bm{99.9}$ \\
    \multirow{3}{*}{\textbf{+DistMult}} & \xmark & \cmark & $83.5$ & $67.1$ & $\bm{100.0}$ & $\bm{100.0}$ & $42.4$ & $25.2$ & $44.1$ & $98.7$ \\
    & \cmark & \xmark & $84.0$ & $68.0$ & $\bm{100.0}$ & $\bm{100.0}$ & $35.2$ & $11.6$ & $41.8$ & $99.8$ \\
    & \cmark & \cmark & $\bm{84.5}$ & $\bm{68.9}$ & $\bm{100.0}$ & $\bm{100.0}$ & $\bm{59.3}$ & $\bm{49.3}$ & $\bm{58.8}$ & $94.4$ \\
    \bottomrule
    \end{tabular}
    \caption{Link prediction results on \VisualSem{}'s validation set with additional features. We show best overall scores per metric in bold, and we underline best scores for a single model across all features per metric.}
    \label{tab:graph-results-additional-features-validation}
\end{table*}

\section{Node and Edge Gating}\label{sec:appendix:node_and_edge_gating}
\subsection{Node Gating}\label{sec:appendix:node_gating}
We gate features $\{\bm{t}_i, \bm{m}_i\}$ directly with node embeddings $\bm{v}_i$, and denote the node gating function $f_n$ with parameters $\theta_n$ to transform node embeddings and additional features and compute \textit{informed} node embeddings.

\paragraph{Text}
When integrating only textual features into node embeddings $\bm{v}_i$, we compute $\bm{v}^t_i = f_n( \bm{v}_i, \bm{t}_i )$ as below.
\begin{align}
    s_g^t &= \text{MLP}_g^t([\bm{v}_i; \bm{W}_g^t \cdot \bm{t}_i]), \notag\\
    \bm{v}^t_i &= s_g^t \cdot \bm{t}_i + (1-s_g^t) \cdot \bm{v}_i , \label{eqn:node-gating-node-with-gloss}
\end{align}
\noindent
where $\bm{W}_g^t$ is a trained projection matrix for gloss features, $s_g^t$ is a gating scalar computed using a multi-layer perceptron MLP$_g^t$, and $\bm{v}^t_i$ are node embeddings informed by glosses $\mathcal{T}_i$.

\paragraph{Image}
When integrating only image features into node embeddings  $\bm{v}_i$, we compute $\bm{v}^m_i = f_n( \bm{v}_i, \bm{m}_i )$ as below.
\begin{align}
    s_g^m &= \text{MLP}_g^m([\bm{v}_i; \bm{W}_g^m \cdot \bm{m}_i]), \notag\\
    \bm{v}^m_i &= s_g^m \cdot \bm{m}_i + (1-s_g^m) \cdot \bm{v}_i , \label{eqn:node-gating-node-with-image}
\end{align}
\noindent
where similarly $\bm{W}_g^m$ is a trained projection matrix for image features, $s_g^m$ is a gating scalar computed with MLP$_g^m$, and $\bm{v}^m_i$ are node embeddings informed by images $\mathcal{I}_i$.

\paragraph{Text+Image}
Finally, when integrating both textual and image features into $\bm{v}_i$, informed node embeddings $\bm{v}^{t,m}_i = f_n( \bm{v}_i, \bm{t}_i, \bm{m}_i )$ are computed below.
\begin{align}
    \bm{v}_i^{t,m} &= \bm{W}_g^{t,m} \cdot [\bm{v}_i^t; \bm{v}_i^m] \label{eqn:node-gating-node-with-gloss-and-image}
\end{align}
\noindent
where $\bm{W}_g^{t,m}$ is a projection matrix and $\bm{v}^{t,m}_i$ are node embeddings informed by both glosses and images.

\subsection{Edge (Relation) Gating}\label{sec:appendix:edge_gating}
Given a triplet $(v_i, e_r, v_j)$, we gate text features $\bm{t}_i$ with edge embeddings $\bm{e}_r$ as below.
\begin{align}
    \bm{t}_r &= \frac{\bm{W}_d^t \bm{t}_i + \bm{W}_d^t \bm{t}_j}{2}, \notag\\
    s_d^t &= \text{MLP}_d^t( [\bm{e}_r; \bm{t}_r]), \notag\\
    \bm{e}^t_{r} &= s_d^t \cdot \bm{t}_r + (1-s_d^t) \cdot \bm{e}_r, \label{eqn:edge-gating-edge-with-gloss}
\end{align}
\noindent
where $\bm{W}_d^t$ is a trained projection matrix, $\bm{t}_r$ are average gloss features for nodes $v_i$ and $v_j$, $\bm{e}^t_{r}$ are edge embeddings informed by glosses $\mathcal{T}_i \cup \mathcal{T}_j$, and $s_d^t$ is a gating scalar computed using a multi-layer perceptron MLP$_d^t$.
Similarly, we gate image features $\bm{m}_i$ with edge embeddings $\bm{e}_r$ as below.
\begin{align}
    \bm{m}_r &= \frac{\bm{W}_d^m \bm{m}_i + \bm{W}_d^m \bm{m}_j}{2}  \notag\\
    s_d^m &= \text{MLP}_d^m( [\bm{e}_r; \bm{m}_r ]), \notag\\
    \bm{e}^m_{r} &= s_d^m \cdot \bm{m}_r + (1-s_d^m) \cdot \bm{e}_r, \label{eqn:edge-gating-edge-with-image}
\end{align}
\noindent
where $\bm{W}_d^m$ is a trained projection matrix, $\bm{m}_r$ are average image features for nodes $v_i$ and $v_j$, $\bm{e}^m_{r}$ are edge embeddings informed by images $\mathcal{I}_i \cup \mathcal{I}_j$, and $s_d^m$ is a gating scalar computed using a multi-layer perceptron MLP$_d^m$.
Finally, we combine both text and image features as described in Equation \ref{eqn:edge-gating-edge-with-gloss-and-image}.
\begin{align}
    \bm{e}_{r}^{t,m} &= \bm{W}_d^{t,m} [\bm{e}_{r}^t; \bm{e}_{r}^m] \label{eqn:edge-gating-edge-with-gloss-and-image}
\end{align}
\noindent
where $\bm{W}_d^{t,m}$ is a parameter matrix and $\bm{e}_{i,j}^{r}$ are edge embeddings informed by glosses and images.

\subsection{Training}\label{sec:appendix:training}
To train hybrid models GAT+DistMult and GraphSage+DistMult with additional features, we use
(1) only gloss features $\mathcal{T}_i$, i.e. node and edge embeddings computed by Equations \ref{eqn:node-gating-node-with-gloss} and \ref{eqn:edge-gating-edge-with-gloss}, respectively;
(2) only image features $\mathcal{I}_i$, i.e. node and edge embeddings computed by Equations \ref{eqn:node-gating-node-with-image} and \ref{eqn:edge-gating-edge-with-image}, respectively;
(3) both gloss and image features $\mathcal{T}_i$, $\mathcal{I}_i$, i.e. node and edge embeddings computed by Equations \ref{eqn:node-gating-node-with-gloss-and-image} and \ref{eqn:edge-gating-edge-with-gloss-and-image}, respectively.

\section{Downstream Tasks}

\subsection{NER Hyperparameters}

We use a BERT base architecture for the German NER model (GermEval) and a BERT large architecture for the English NER model (WNUT-17), in both cases using case-sensitive models.
For both GermEval and WNUT-17, the best performing model configuration uses a projection of the original \VisualSem{} graph representations (100-dimensional) into the respective BERT encoder dimensions (mBERT 768-dimensional, English monolingual BERT 1024-dimensional).
These projected features are concatenated with the corresponding BERT encoding, which is then used as input to the prediction head.

\subsection{NER Ablation}

We report additional NER experiment results on the WNUT-17 and GermEval validation sets using (1) concatenation with $100 \times d \in \{1024, 768\}$ projection matrix (Table \ref{tab:NER-dev-results-1}),
and on the WNUT-17 validation set using gating with $100 \times 1024$ projection matrix (Table \ref{tab:NER-dev-results-2}), gating with $100 \times 100$ projection matrix (Table \ref{tab:NER-dev-results-3}), concatenation with $100 \times 100$ projection matrix (Table \ref{tab:NER-dev-results-4}),gating without projection plus fine-tuning node hidden states (Table \ref{tab:NER-dev-results-5}), and concatenation without projection plus fine-tuning node hidden states (Table \ref{tab:NER-dev-results-6}).

\begin{table}[t!]
    \centering
    \small
    \begin{tabular}{l l ccc}
    \toprule
    & &  \textbf{Precision} & \textbf{Recall} & \textbf{F1 Score} \\
    \midrule
     \parbox[t]{2mm}{\multirow{6}{*}{\rotatebox[origin=c]{90}{\textbf{EN}}}}
     &{\textbf{Baseline}} &  $\bm{70.1}$ & $50.7$ & $58.8$  \\
    \cmidrule{2-5}
    & \textbf{+}$\bm{h}_i^\text{\bf NODE}$ & $68.7$ & $53.1$ & $59.9$ \\
     & \textbf{+}$\bm{h}_i^\text{\bf IMG}$ &  $68.5$ & $\bm{54.4}$ & $\bm{60.7}$  \\
     & \textbf{+}$\bm{h}_i^\text{\bf TXT}$  &$65.8$ & $\bm{54.4}$ & $59.6$  \\
     & \textbf{+}$\bm{h}_i^\text{\bf TXT+IMG}$ &$64.2$ & $53.0$ & $58.1$\\
    \midrule
    \parbox[t]{2mm}{\multirow{6}{*}{\rotatebox[origin=c]{90}{\textbf{DE}}}}
    &{\textbf{Baseline}} & $84.7$ & $87.8$ & $86.2$             \\
    \cmidrule{2-5}
     &\textbf{+}$\bm{h}_i^\text{\bf NODE}$  & $\bm{85.2}$          & $\bm{88.5}$          & $\bm{86.8}$   \\
     & \textbf{+}$\bm{h}_i^\text{\bf IMG}$ &$84.8$          & $88.1$          & $86.4$  \\
     & \textbf{+}$\bm{h}_i^\text{\bf TXT}$  &$85.0$         & $88.1$          & $86.5$  \\
     & \textbf{+}$\bm{h}_i^\text{\bf TXT+IMG}$ &$85.0$          & $88.3$          & $86.6$ \\
    \bottomrule
    \end{tabular}
    \caption{Concatenation with $100 \times d \in \{1024, 768\}$ projection matrix}
    \label{tab:NER-dev-results-1}
\end{table}

\begin{table}[t!]
    \centering
    \small
    \begin{tabular}{l ccc}
    \toprule
    &  \textbf{Precision} & \textbf{Recall} & \textbf{F1 Score} \\
    \midrule 
    {\textbf{Baseline}} &  $\bm{70.1}$ & $50.7$ & $58.5$  \\
    \midrule
    \textbf{+}$\bm{h}_i^\text{\bf NODE}$ & $66.3$   & $50.7$ & $57.5$  \\
    \textbf{+}$\bm{h}_i^\text{\bf IMG}$ &  $58.6$  & $51.3$  & $54.7$ \\
    \textbf{+}$\bm{h}_i^\text{\bf TXT}$ & $68.0$          & $\bm{53.5}$          & $\bm{59.9}$ \\
    \textbf{+}$\bm{h}_i^\text{\bf TXT+IMG}$ & $65.7$          & $52.5$          & $58.4$      \\
    \bottomrule
    \end{tabular}
    \caption{Gating with $100 \times 1024$ projection matrix}
    \label{tab:NER-dev-results-2}
\end{table}

\begin{table}[t!]
    \centering
    \small
    \begin{tabular}{l ccc}
    \toprule
    &  \textbf{Precision} & \textbf{Recall} & \textbf{F1 Score} \\
    \midrule 
    {\textbf{Baseline}} & $\bm{70.1}$          & $50.7$ & $58.5$          \\
    \midrule
    \textbf{+}$\bm{h}_i^\text{\bf NODE}$ & $65.5$ & $\bm{53.2}$ & $58.7$ \\
     \textbf{+}$\bm{h}_i^\text{\bf IMG}$ &$68.6$ & $51.6$ & $\bm{58.9}$ \\
    \textbf{+}$\bm{h}_i^\text{\bf TXT}$ & $67.3$ & $50.0$ & $57.4$ \\
     \textbf{+}$\bm{h}_i^\text{\bf TXT+IMG}$ & $67.9$ & $51.2$ & $58.4$ \\
    \bottomrule
    \end{tabular}
    \caption{Gating with $100 \times 100$ projection matrix}
    \label{tab:NER-dev-results-3}
\end{table}

\begin{table}[t!]
    \centering
    \small
    \begin{tabular}{l ccc}
    \toprule
    &  \textbf{Precision} & \textbf{Recall} & \textbf{F1 Score} \\
    \midrule 
    {\textbf{Baseline}} & $\bm{70.1}$ & $50.7$ & $58.5$ \\
    \midrule
    \textbf{+}$\bm{h}_i^\text{\bf NODE}$ & $54.7$  & $45.9$ & $49.9$ \\
     \textbf{+}$\bm{h}_i^\text{\bf IMG}$ & $66.3$ & $52.8$ & $\bm{58.8}$ \\
    \textbf{+}$\bm{h}_i^\text{\bf TXT}$ & $65.9$ & $\bm{53.0}$ & $\bm{58.8}$ \\
     \textbf{+}$\bm{h}_i^\text{\bf TXT+IMG}$ & $67.8$ & $51.3$ & $58.4$\\
    \bottomrule
    \end{tabular}
    \caption{Concatenation with $100 \times 100$ projection matrix}
    \label{tab:NER-dev-results-4}
\end{table}

\begin{table}[t!]
    \centering
    \small
    \begin{tabular}{l ccc}
    \toprule
    &  \textbf{Precision} & \textbf{Recall} & \textbf{F1 Score} \\
    \midrule 
    {\textbf{Baseline}} & $70.1$ & $50.7$ & $58.5$ \\
    \midrule
    \textbf{+}$\bm{h}_i^\text{\bf NODE}$ & $\bm{70.7}$ & $51.7$ & $59.7$ \\
     \textbf{+}$\bm{h}_i^\text{\bf IMG}$ & $67.1$ & $50.8$ & $57.9$\\
    \textbf{+}$\bm{h}_i^\text{\bf TXT}$ & $69.8$ & $\bm{52.9}$ & $\bm{60.2}$ \\
     \textbf{+}$\bm{h}_i^\text{\bf TXT+IMG}$ & $55.2$ & $48.1$ & $51.4$ \\
    \bottomrule
    \end{tabular}
    \caption{Gating without projection plus fine-tuning node hidden states}
    \label{tab:NER-dev-results-5}
\end{table}

\begin{table}[t!]
    \centering
    \small
    \begin{tabular}{l ccc}
    \toprule
    &  \textbf{Precision} & \textbf{Recall} & \textbf{F1 Score} \\
    \midrule 
    {\textbf{Baseline}} & $\bm{70.1}$ & $50.7$ & $58.5$ \\
    \midrule
    \textbf{+}$\bm{h}_i^\text{\bf NODE}$ & $67.7$ & $51.2$ & $58.3$ \\
     \textbf{+}$\bm{h}_i^\text{\bf IMG}$ & $64.6$ & $49.8$ & $56.2$ \\
    \textbf{+}$\bm{h}_i^\text{\bf TXT}$ & $65.5$ & $\bm{52.3}$ & $58.2$ \\
     \textbf{+}$\bm{h}_i^\text{\bf TXT+IMG}$ & $67.4$ & $52.0$ & $\bm{58.7}$ \\
    \bottomrule
    \end{tabular}
    \caption{Concatenation without projection plus fine-tuning node hidden states}
    \label{tab:NER-dev-results-6}
\end{table}

\subsection{Visual Verb Sense Disambiguation Ablation}
For the German portion of the datasets, there are 707 tuples in the form \textit{English verb, English query, German verb}. We divided the dataset into $75\%$ training, $10\%$ validation and $15\%$ test splits, and used minibatch size $16$, following the set up in \citet{gella2019crosslingual}. We use learning rate $0.0005$ and dropout with probability $0.1$ across all settings.

Table \ref{tab:Multisense-dev-results} reports validation set results on MultiSense using different graph representations.

\begin{table}[t!]
    \centering
    \small
    \begin{tabular}{rrrrr}
    \toprule
    {\textbf{Our Baseline}} & \textbf{+}$\bm{h}_i^\text{\bf NODE}$ &\textbf{+}$\bm{h}_i^\text{\bf TXT}$ & \textbf{+}$\bm{h}_i^\text{\bf TXT+IMG}$ & \textbf{+}$\bm{h}_i^\text{\bf IMG}$ \\
     \midrule
     95.7 & 96.1 & \textbf{97.8} & 96.4 & 95.5 \\
    \bottomrule
    \end{tabular}
    \caption{Accuracy on the German MultiSense validation set.}
    \label{tab:Multisense-dev-results}
\end{table}

\end{document}